\documentclass[runningheads,a4paper]{llncs}

\usepackage{amssymb}
\usepackage{subfigure}
\usepackage{graphicx}
\usepackage{booktabs}
\usepackage{tabularx}
\usepackage{multirow}
\usepackage{rotating}
\usepackage{cite}

\usepackage{url}
\urldef{\mailsa}\path|{alfred.hofmann, ursula.barth, ingrid.haas, frank.holzwarth,|
\urldef{\mailsb}\path|anna.kramer, leonie.kunz, christine.reiss, nicole.sator,|
\urldef{\mailsc}\path|erika.siebert-cole, peter.strasser, lncs}@springer.com|
\urldef{\mymail}\path|xirong.li@gmail.com|

\usepackage{array}
\newcolumntype{L}{>{\arraybackslash}m{0.2\columnwidth}}

\newcommand{\keywords}[1]{\par\addvspace\baselineskip
\noindent\keywordname\enspace\ignorespaces#1}

\usepackage{verbatim}

\usepackage{colortbl}

\definecolor{mygray}{gray}{.75}

\begin{document}

\mainmatter  

\title{Adaptive Tag Selection for Image Annotation}

\titlerunning{Adaptive Tag Selection for Image Annotation}

%
%
\author{Xixi He\textsuperscript{1,2}, Xirong Li\textsuperscript{1,2,3}\footnote{Corresponding author (\mymail)}, Gang Yang\textsuperscript{1,2}, Jieping Xu\textsuperscript{1,2}, Qin Jin\textsuperscript{1,2}}
\authorrunning{He \emph{et al.}}

\institute{
\textsuperscript{1}Key Lab of DEKE, Renmin University of China, 100872 China \\
\textsuperscript{2}Multimedia Computing Lab, Renmin University of China, 100872 China\\
\textsuperscript{3}Shanghai Key Laboratory of Intelligent Information Processing, 200443 China \\
}

%
%

\toctitle{Lecture Notes in Computer Science}
\tocauthor{Authors' Instructions}
\maketitle

\begin{abstract}
Not all tags are relevant to an image, and the number of relevant tags is image-dependent.
Although many methods have been proposed for image auto-annotation,
the question of how to determine the number of tags to be selected per image remains open.
The main challenge is that for a large tag vocabulary, there is often a lack of ground truth data for acquiring optimal cutoff thresholds per tag.
In contrast to previous works that pre-specify the number of tags to be selected, 
we propose in this paper \textit{adaptive tag selection}.
The key insight is to divide the vocabulary into two disjoint subsets, namely a seen set consisting of tags having ground truth available for optimizing their thresholds and a novel set consisting of tags without any ground truth.
Such a division allows us to estimate how many tags shall be selected from the novel set according to the tags that have been selected from the seen set. 
The effectiveness of the proposed method is justified by our participation in the ImageCLEF 2014 image annotation task.
On a set of 2,065 test images with ground truth available for 207 tags,
the benchmark evaluation shows that
compared to the popular top-$k$ strategy which obtains an F-score of 0.122,
adaptive tag selection achieves a higher F-score of 0.223.
Moreover, by treating the underlying image annotation system as a black box,
the new method can be used as an easy plug-in to boost the performance of existing systems.
\end{abstract}


\section{Introduction} \label{sec:intro}

\begin{figure}[tb!]
\centering
\includegraphics[width=\columnwidth]{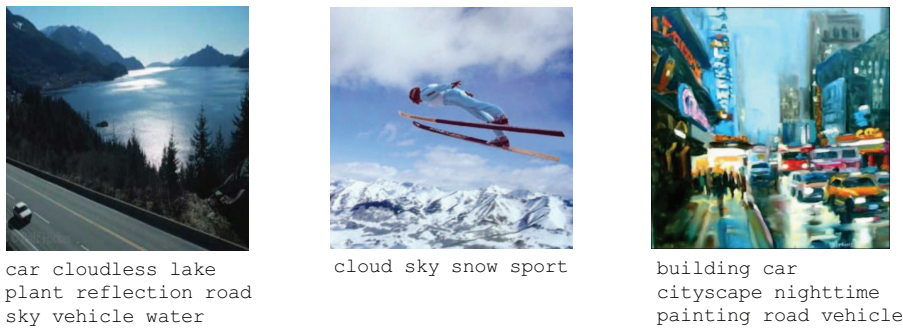}
\caption{\textbf{Internet images and ground truth tags} from the development set of the ImageCLEF 2014 image annotation task \cite{clefanno-2014}.
The fact that the number of relevant tags varies over images motivates us to study how to assign a proper number of tags for annotating unlabeled images.}
\label{fig:image-examples}
\end{figure}

Annotating images by computers is crucial for accessing the many unlabeled images at a semantic level.
Due to the semantic gap, i.e., the lack of correspondence between visual features extracted from the pictorial content and a user's interpretation of the content, image auto-annotation is challenging. 
Labeling arbitrary images on the Internet is even more difficult,
as a relatively simple concept may exhibit significant diversity in its visual appearance. 
The imagery of a concept does not limit to realistic photographs, but can also be artificial correspondences such as posters, drawings, and cartoons,
as exemplified in Fig. \ref{fig:image-examples}.
On the one hand, a large array of tags need to be modeled for depicting the diverse content of Internet image collections,
while on the other hand, as not all tags are relevant to a specific image, we need to make binary assignments of the tags to that image.

Quite a few methods have been proposed for image annotation, either by building visual classifiers per tag \cite{clef2011-koen,mm2013-tagrel} or by propagating tags from visually similar images \cite{ieee2012-wang,icmr2014-ballan}.
Given a novel image and a tag vocabulary, these methods first compute each tag's relevance score with respect to the given image, and sort the tags in descending order by their scores.
The top-$k$ ranked tags are preserved as predicted annotations of the image.
The choice of $k$ reflects the trade-off between precision and recall.
In previous works, a fixed value of $k$ is used for all images, where $k=5$ is a common choice \cite{icmr2014-ballan}.
Notice however that the number of relevant tags varies over images. 
Hence, it is not surprising that such a top-$k$ strategy gives suboptimal results.

A good method for image annotation shall be able to adaptively determine which tags to be selected per image.
Since choosing a proper $k$ per image is difficult, 
one might consider a thresholding strategy that a specific tag is selected if its relevance score is larger than a given threshold.
In \cite{clef2011-koen}, the thresholds are optimized by maximizing a combined metric of precision and recall, say F-score, on training data.
Despite its good performance, optimizing thresholds per tag requires ground-truthed data, 
which indicates the relevance of an image with respect to a given tag, for all tags in consideration.
Consequently, this strategy is inapplicable to novel tags which have no ground truth available.

This paper studies adaptive tag selection for image annotation. 
In particular, 
given a ranked list of tags produced by a specific image annotation system for a test image,
we aim to answer the question of how to adaptively determine a proper number of tags for annotating the test image.
To that end, an adaptive top-$k$ tag selection method is proposed, which beats the standard top-$k$ strategy with ease. 

\section{Related Work} \label{sec:related}

Image annotation, as an important topic in the multimedia field, has been actively studied.
A noticeable effort is to build scalable image annotation systems based on large-scale user-contributed data instead of limited-scale expert-labeled training data \cite{ieee2012-wang,tmm09-xirong,tist2010-tang,semanticfield-tmm12}.
While the number of tags that can be modeled is increasing,
it remains unclear how to select a proper number of tags to annotate an unlabeled image.
This problem is overlooked, because most of the existing works either assess the top-$k$ ranked tags \cite{icmr2014-ballan,ieee2012-wang} or the entire tag ranking list \cite{tmm09-xirong}.

To make the number of selected tags adaptive, several works employ a thresholding strategy by keeping tags that have scores larger than specified thresholds. In \cite{clef2013-cea}, the authors use the sum of the average and the standard derivation of the scores w.r.t. a tag as its threshold.
However, according to our observation, there is a lack of evidence supporting that the score of a relevant tag is indeed larger than this threshold.
To find thresholds that are more related to the image annotation performance,
the authors in \cite{clef2011-koen} find a global threshold for all tags by cross-validation on ground-truthed data. 
Due to the diversity in the many tags and their corresponding models, it is unlikely that one threshold is suitable for all tags.
On the other hand, obtaining optimal threshold for each tag is difficult as this would require full annotations.
Hence, it is worthwhile to study adaptive tag selection given incomplete ground truth, and this has not been well explored in the literature.

\section{Adaptive Tag Selection} \label{sec:method}

As aforementioned, for a given tag vocabulary, optimizing the cutoff thresholds per tag is inapplicable to novel tags.
Nevertheless, we can safely assume that we have access to a set of training images manually  yet incompletely labeled using a subset of the vocabulary.
Consequently, depending on whether a tag has a number of ground-truthed images available,
the vocabulary can be divided into two disjoint subsets,
i.e., the \emph{seen set} consisting of tags with ground truth 
and the \emph{novel set} consisting of tags without ground truth.
We assume tags are uniformly assigned to the two sets,
and consequently for a given image, its relevant tags have the same occurrence probability in the seen set and in the novel set.
Hence, we can estimate the number of tags to be selected from the novel set according to the number of tags that have been selected from the seen set.

To describe the above idea more formally, we introduce some notation.
Let $x$ be an image, $t$ be a tag, and $\mathcal{V}$ be a tag vocabulary.
We use $\mathcal{V}_{seen}$ to denote the seen set, and $\mathcal{V}_{novel}$ for the novel set.
Let $\mathcal{X}_{train}$ be a set of training images which are manually labeled using $\mathcal{V}_{seen}$ only.
In order to select from $\mathcal{V}$ relevant tags to annotate $x$,
we need an image tag relevance function $f(x,t)$ which computes the relevance score of $t$ with respect to $x$.
The popular top-$k$ strategy annotates $x$ by sorting $\mathcal{V}$ in descending order by $f(x,t)$ and selecting the top $k$ ranked tags.
In contrast to previous works which designate $k$ in advance,
we make $k$ variable by selecting tags from $\mathcal{V}_{novel}$ based on the selection on $\mathcal{V}_{seen}$.
Given a specific test image, let $\mathcal{A}$ be the tags that have been selected from $\mathcal{V}_{seen}$.
Based on our hypothesis that relevant tags of the test image have the same occurrence probability in $\mathcal{V}_{seen}$ and $\mathcal{V}_{novel}$,
we propose to estimate the number of tags to be selected from $\mathcal{V}_{novel}$ as 
\begin{equation} \label{eq:k_novel}
k_{novel} := |\mathcal{V}_{novel}| \cdot \frac{|\mathcal{A}|}{|\mathcal{V}_{seen}|},
\end{equation}
where $|\cdot|$ returns the set cardinality. 
Concerning $\mathcal{A}$, we obtain it by thresholding:
\begin{equation} \label{eq:A}
\mathcal{A} := \{t \in \mathcal{V}_{seen} | f(x,t) > \tau_t\},
\end{equation}
where $\tau_t$ is the corresponding threshold found by maximizing the tag's F-score on $\mathcal{X}_{train}$.
Since $\mathcal{A}$ is image dependent, the proposed method will select a variable number of tags.
The number of selected tags is $|\mathcal{A}| + k_{novel}$. In a rare case where $\mathcal{A}$ is empty, we switch back to the top-$k$ strategy.

Further, as $\mathcal{A}$ is constructed based on the learned thresholds, 
we consider refining the relevance scores for $\mathcal{V}_{novel}$ by exploiting $\mathcal{A}$ as pseudo labels.
Tags that are semantically close to $\mathcal{A}$ shall be strengthened,
and in the meanwhile tags from $\mathcal{A}$ that are more reliable shall have more weights.
We implement this thought by updating the relevance score of $t \in \mathcal{V}_{novel}$ as
\begin{equation} \label{eq:update}
f(x,t) \leftarrow w \cdot f(x,t) + (1-w) \frac{1}{|\mathcal{A}|} \sum_{t' \in \mathcal{A}} sim(t,t') \cdot (\frac{f(x,t')}{\tau_{t'}} - 1),
\end{equation}
where $w$ is a weighting parameter, and $sim(t,t')$ measures semantic similarity between two tags.
In Eq. (\ref{eq:update}), $f(x,t')$ is divided by $\tau_{t'}$ as an effect of scale normalization.
We compute $sim(t,t')$ using the Flickr Context Similarity \cite{mm09-jiang}, which is based on the Normalized Google Distance \cite{tkde04-ngd},
but with tag statistics acquired from Flickr image collections instead of Google indexed web pages.

Notice that the proposed tag selection method treats the underlying image annotation system as a black box.
Hence, it can be easily used as a plug-in to boost the performance of existing methods.

\section{Image Annotation System} \label{sec:system}

As we aim for modeling many tags, we build an image annotation system with its classifiers trained purely on web data with no need of extra manual labeling. 
This property makes the system more scalable with respect to the number of tags compared to systems relying on manually labeled data.
The main components of the system, namely visual features, training data, and image annotation models, are depicted as follows.

\textbf{Visual features}. For each image, we extract a bag of visual words using the color descriptor software \cite{koen-tpami10}.
A precomputed codebook of size 4,000 is used to quantize densely sampled SIFT descriptors.
To improve the spatial discriminativeness of the feature, we further consider 1x1+1x3 spatial pyramids.
This results in a visual feature vector of 16,000 dimensions per image.

\textbf{Training data acquisition}. 
We leverage three sources of training data, all of which were acquired with manual annotation for free.
The first set is a set of 250K images \cite{clefanno-2013}, collected by querying web image search engines.
The second set contains one million images with user-click count, released by the MSR Bing \cite{mm13-hua}.
The third set consists of four million user-tagged images from Flickr.
As the training sets come from different sources with different (noisy) annotation information,
we describe how to select positive training examples for a specific tag $t$ from the individual sets.

For the 250K web images, as they were collected from three web image search engines, namely Google, Yahoo, and Bing,
each image $x$ can be described by a triplet $<q, r, s>$, where $q$ represent a query tag, $r$ is the rank of $x$ in the search results of $q$ returned by an specific search engine $s$.
Because a given image might be retrieved by different queries or by the same query but with different search engines,
it can be associated with multiple triplets, denoted as $<q_i, r_i, s_i>$, $i=1,\ldots,l$, where $l$ is the number of triplets.
To estimate the relevance of $x$ with respect to $t$, we propose to compute a search engine based score as
\begin{equation} \label{eq:search-score}
relevance_{search}(x,t)=\sum_{i=1}^l \delta(q_i, t) \frac{w(s_i)}{\sqrt{r_i}},
\end{equation}
where $\delta(q_i,t)$ returns 1 if $q_i$ and $t$ are the same, and 0 otherwise.
The variable $w(s_i)$ indicates the weight of a specific search engine,
which is empirically set to be 1, 0.5, and 0.5 for Google, Yahoo, and Bing, respectively.

For the user-clicked set, each image is associated with a textual query and the accumulated count of user clicks.
A larger click count indicates that the image is more likely to be relevant to the query \cite{mm13-hua}.
We thus match $t$ with queries and use the corresponding click count as the relevance score.

For the Flickr set, we compute tag relevance scores using the semantic field method \cite{semanticfield-tmm12},
which is computationally more efficient than visual based approaches \cite{tmm09-xirong}.
Given an image with its user tags including $t$, the semantic field method estimates the relevance of $t$ to the image 
by considering the semantic similarity between $t$ and the other tags. 
We again use the Flickr Context Similarity to measure the tag-wise similarity.

For the given tag, we obtain its positive training examples from each of the three sets by sorting images in descending order by their relevance scores and preserve the top 1,000 ranked images.

\textbf{Annotation models}. For each tag we instantiate its $f(x,t)$ by learning two-class SVM classifiers from the three training sets separately. As the training data is overwhelmed by negative examples, 
we train classifiers by the Negative Bootstrap algorithm \cite{relneg-tmm13}. 
Different from sampling negative examples at random, Negative Bootstrap iteratively selects negative examples which are most misclassified by present classifiers, and thus most relevant to improve classification.
Per iteration, the algorithm randomly samples 10$\times$1,000=10,000 examples to form a candidate set.
An ensemble of classifiers obtained in the previous iterations are used to classify each candidate example.
The top 1,000 most misclassified examples are selected and used together with the 1,000 positives to train a new classifier.
For the consideration of efficiency, we use Fast intersection kernel SVMs (FikSVM) \cite{cvpr08-fiksvm}.
For each of the three sets, we conduct Negative Bootstrap with 10 iterations, producing in total 3$\times$10 FikSVMs per tag.
These FikSVMs are further compressed into a single model such that the annotation time complexity depends only on the feature dimensionality.

We observe that models trained on the three sets are complementary to each other to some extent. 
We therefore combine the models in a linear late fusion manner.
Our previous study shows that weights optimized by coordinate ascent consistently outperforms averaging \cite{ruc-clef2013}.
So we continue this good practice, and learn the fusion weights by coordinate ascent on $\mathcal{X}_{train}$.

\section{Experiments} \label{sec:exp}

\subsection{Experimental setup} \label{ssec:runs}

To verify the effectiveness of the proposed method, we participated in the ImageCLEF 2014 image annotation task \cite{clefanno-2014}, a benchmark for developing scalable image annotation systems without using manually labeled training examples.
The task asks the participated systems to annotate unlabeled test images using a vocabulary of 207 tags, see the Appendix.
There are 2,065 test images manually labeled using the vocabulary.
Notice that the ground truth of the test set is unavailable to the participants,
so the result reported in this paper are from the official evaluation\footnote{\url{http://imageclef.org/2014/annotation/results}. 
We ignore test images which do not have full ground truth with respect to the 207 tags, so our numbers differ from the original results.} 
and extra evaluation provided by the organizers on our request.
A development set of 1,000 labeled images are provided for 107 tags, whilst no ground truth is given for the remaining 100 tags.
This setting allows us to evaluate the viability of the proposed method.

\textbf{Baselines}. In addition to the common top-5 strategy, 
we compare with \cite{clef2013-cea}, which computes for each tag the average ($\mu$) and the standard deviation ($\sigma$) of the scores, 
and selects the tag having a score above $\mu + \sigma$. 
Since the development set allows us to find optimal thresholds for each tag in $\mathcal{V}_{seen}$, 
we also try to reconstruct the thresholds by linear combination of $\mu$ and $\sigma$ with the tag-independent coefficients solved by least square fitting. The threshold of $t \in \mathcal{V}_{novel}$ is estimated by linearly combining  $\mu$ and $\sigma$  with the learned coefficients.
We denote this strategy as lsq($\mu,\sigma$).
Notice that for a fair comparison, all methods are given the same tag rank lists produced by the system described in Section \ref{sec:system}.

\textbf{Performance metrics}. 
We report mean F-score (mF) and mean Average Precision (mAP) at the image level, 
which measures the quality of the selected tags and the quality of the entire tag ranking list, respectively.

\subsection{Results} \label{ssec:results}

As shown in Table \ref{tab:compare}, the proposed method clearly outperforms the other methods for tag selection.
In order to reveal if the gain is mainly contributed by selecting tags from $\mathcal{V}_{seen}$ using the learned thresholds,
for both $\mu + \sigma$ and lsq($\mu, \sigma$) methods,
we use the learned thresholds as an alternative to the predicted thresholds for $\mathcal{V}_{seen}$.
Though their mF-score increases as shown in the fifth and sixth rows in Table \ref{tab:compare},
the proposed method maintains the leading position.

\begin{table}[tb!]
\renewcommand{\arraystretch}{1.3}
\caption{\textbf{Performance of different methods for tag selection}. 
As the methods are given the same tag rankings, they have the same mAP score of 0.151.}
\label{tab:compare}    
\centering
\scalebox{1}{
\begin{tabular}{@{}l r@{}}
\toprule
Method & mF-score \\
\cmidrule{1-2}
top-5  & 0.122 \\
$\mu+\sigma$ \cite{clef2013-cea} & 0.127 \\
lsq($\mu, \sigma$) & 0.108 \\
learned $\tau$ for $\mathcal{V}_{seen}$, $\mu+\sigma$ for $\mathcal{V}_{novel}$ & 0.153 \\
learned $\tau$ for $\mathcal{V}_{seen}$, lsq($\mu, \sigma$) for $\mathcal{V}_{novel}$ & 0.150 \\
\emph{proposed method} & \textbf{0.223} \\
\bottomrule
\end{tabular}
}
\end{table}

\begin{figure}[tb!]
\centering
 \subfigure[] {
\noindent\includegraphics[width=\columnwidth]{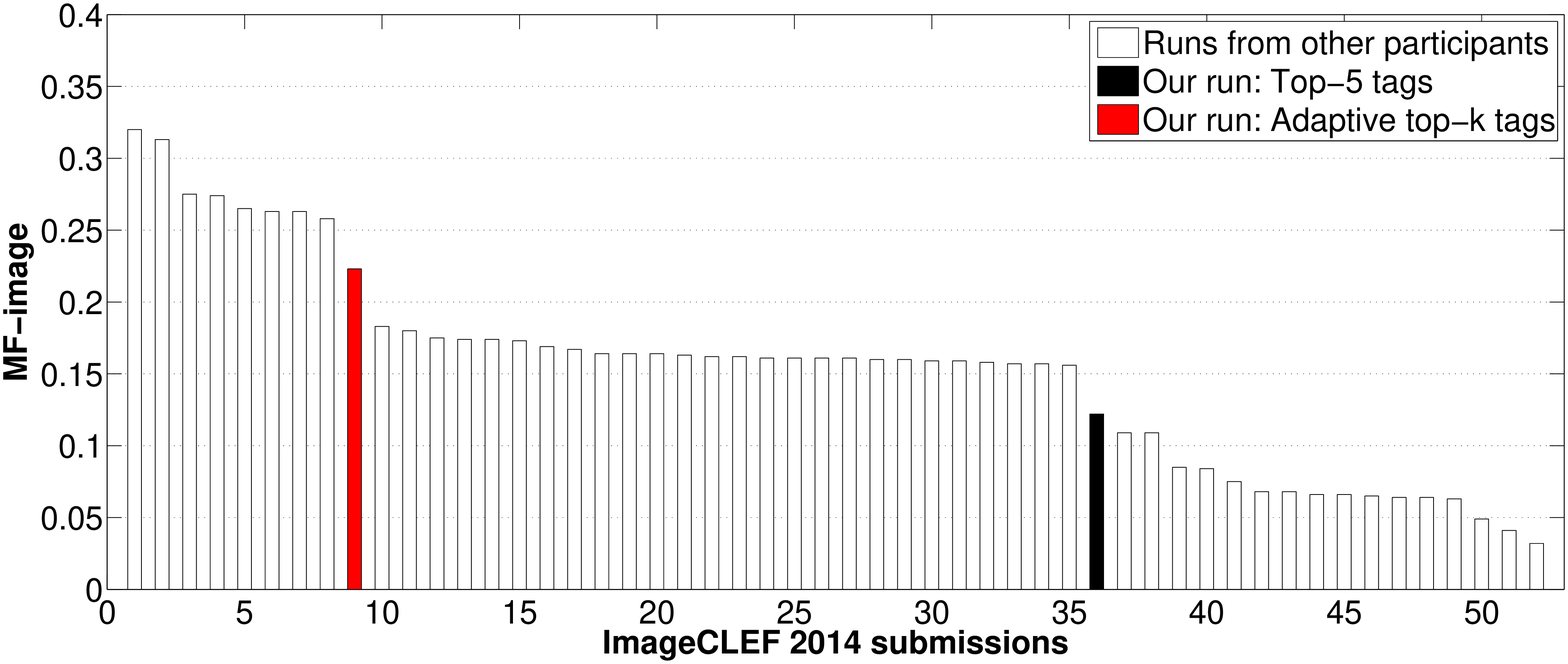}
\label{fig:res-mf}}
 \subfigure[] {
\noindent\includegraphics[width=\columnwidth]{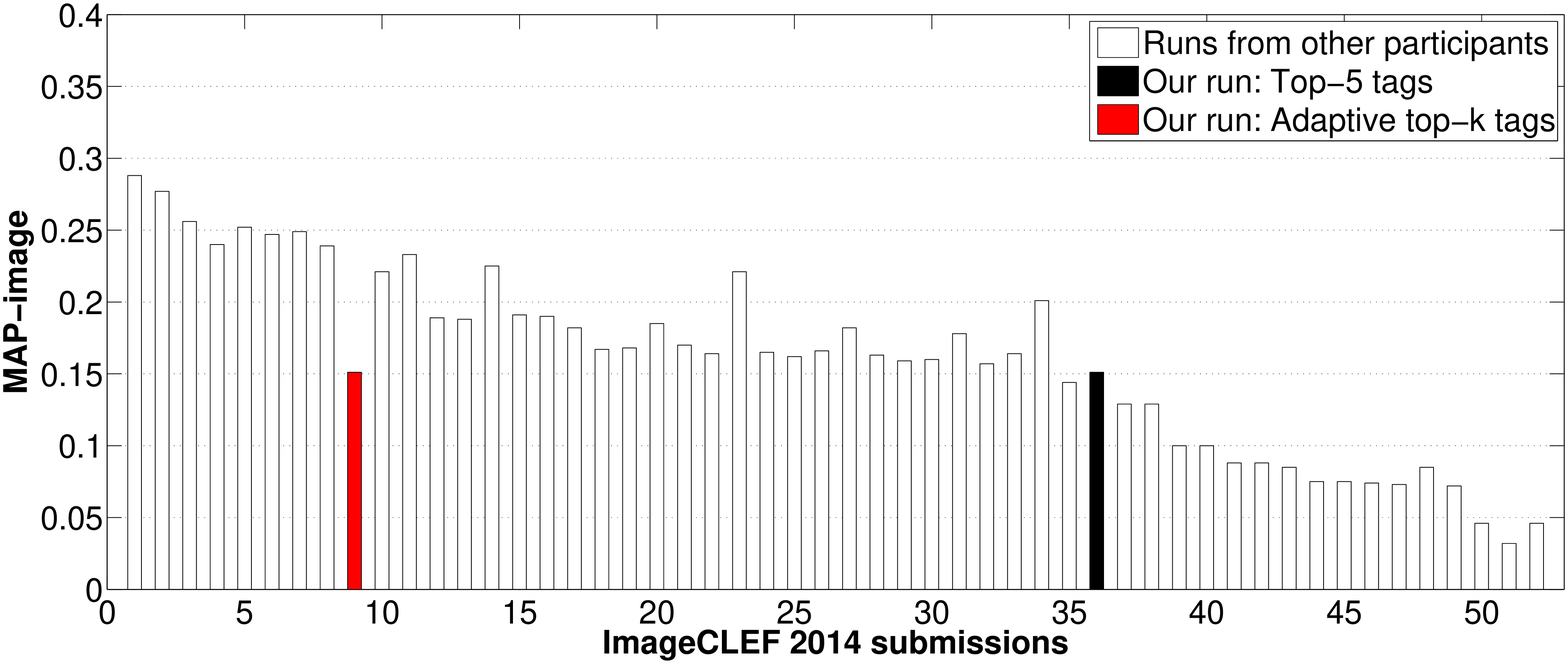}
\label{fig:res-map}}
\caption{
\textbf{Comparing with the ImageCLEF 2014 submissions}.
The submissions are sorted in descending order in terms of the mF scores, as shown in Fig. \ref{fig:res-mf},
with the corresponding mAP scores given in Fi.g \ref{fig:res-map}. 
The inconsistency between the two ranks shows the importance of tag selection for making final annotations.
} \label{fig:clef2014runs}
\end{figure}

Fig. \ref{fig:clef2014runs} shows the performance of our system in the context of all submissions by the 11 teams participated in the ImageCLEF 2014 task.
Even though our submission is ranked 34 out of the 52 submissions in terms of mAP, 
adaptive tag selection brings us to the 9th position in terms of mF.
This result shows the importance of top-$k$ tag selection and the power of the proposed method for top-$k$ tag selection.

For a more intuitive understanding, we present several machine tagging results in Table \ref{tab:anno-results}.

\section{Conclusions} \label{sec:conc}

This paper introduces adaptive tag selection, a new function to enable an image annotation system to label images with a variable number of tags, rather than a fixed number of tags as commonly done in previous works. 
On the base of our experiments in the ImageCLEF 2014 image annotation task,
we offer the following conclusions.
Adaptive tag selection is important when performing binary assignments of tags to individual images.
Given the same image annotation system, with and without adaptive tag selection make a clear difference in annotation performance measured in terms of F-score. The proposed method is found to be effective for adaptive tag selection.

{\small
\subsubsection*{Acknowledgments.}

The authors are grateful to the ImageCLEF coordinators especially dr. Mauricio Villegas for helping evaluate our results.
This research was supported by 
the Fundamental Research Funds for the Central Universities and the Research Funds of Renmin University of China (No. 14XNLQ01), 
NSFC (No. 61303184), SRFDP (No. 20130004120006), BJNSF (No. 4142029),
SRF for ROCS, SEM, 
and Shanghai Key Laboratory of Intelligent Information Processing, China (Grant No. IIPL-2014-002).
}

\section*{Appendix} \label{sec:append}


{\small

\textbf{ImageCLEF 2014 annotation vocabulary}.
The vocabulary consists of 107 dev tags for which we have access to a ground truth set of 1,000 images, 
and 100 novel tags with no ground truth available.\\
\textit{The 107 dev tags} as $\mathcal{V}_{seen}$: 
 aerial airplane baby beach bicycle bird boat book bottle bridge building bus car cartoon castle cat chair child church cityscape closeup cloud cloudless coast countryside daytime desert diagram dog drink drum elder embroidery female fire firework fish flower fog food footwear forest furniture garden grass guitar harbor hat helicopter highway horse indoor instrument lake lightning logo male monument moon motorcycle mountain newspaper nighttime outdoor overcast painting park person phone plant portrait poster protest rain rainbow reflection river road sand sculpture sea shadow sign silhouette sky smoke snow soil space spectacle sport sun sunset table teenager toy traffic train tree tricycle truck underwater unpaved vehicle violin wagon water \\
\textit{The 100 novel tags} as $\mathcal{V}_{novel}$: 
 antelope apple arthropod asparagus avocado banana bear berry blood branch bread broccoli buffalo butterfly camel canidae captive carrot cauliflower cervidae cheese cheetah chimpanzee corn crocodile cucumber donkey egg eggplant elephant equidae felidae flamingo fox fried fruit galaxy giraffe gorilla grape hippopotamus human hunting kangaroo knife koala leaf leopard lettuce lion mammal marsupial meat monkey mud mushroom nebula onion orange ostrich pan pasta pear penguin pig pineapple pinniped pool potato pumpkin rabbit raccoon reptile rhino rice rifle roasted rock rodent sausage soup spider spoon squirrel strawberry submarine tiger tomato trunk tuber turtle vegetable walrus warthog watermelon wild wolf yam zebra zoo \\ [5pt]
\textbf{Performance metrics}. \\
\textit{F-image.} Given a test image $x$, its relevant tag set $R_x$, and a predicted tag set $P_x$, its F-image score is computed as
\begin{equation}
\mbox{F-image}(x)=\frac{2*\mbox{precision}(x)*\mbox{recall}(x)}{\mbox{precision}(x)+\mbox{recall}(x)},
\end{equation}
where $\mbox{precision}(x)$ is $|R_x \cap P_x|/|P_x|$, and $\mbox{recall}(x)$ is $|R_x \cap P_x|/|R_x|$.
Consequently, MF-image is obtained by averaging F-image scores of all test images. \\
\textit{AP-image.}
Given a test image $x$ with $m$ tags sorted in descending order by predicted scores,
its AP-image score is computed as
\begin{equation}
\mbox{AP-image}(x)=\frac{1}{|R_x|}\sum_{i=1}^m \frac{r_i}{i} \delta(i),
\end{equation}
where $r_i$ is the number of relevant tags among the top $i$ tags, and $\delta(i)$ is 1 if the $i$-th tag is in $R_x$, 0 otherwise.
MAP-image is obtained by averaging AP-image scores of all test images.

}

\begin{table} [tb!]
\renewcommand{\arraystretch}{1.2}
\caption{\textbf{Tagging results produced by our image annotation system}.
Tags uniquely selected by the adaptive tag selection method are shown in an {\color{blue}\textit{italic}} font.}
\label{tab:anno-results}
\centering
\scalebox{0.98}{
\begin{tabular}{@{}l >{\arraybackslash}m{3.5in} @{}}
\toprule

 Test image & Predicted tags \\
\cmidrule{1-1} \cmidrule{2-2}

Good results: \\
\multirow{3}{*}{\includegraphics[width=70pt]{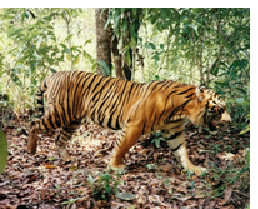}}  & image-id: D1\_8UEAb9SNy2krb \\
& \textbf{Top-5}: tiger forest felidae mammal wolf \\
& \textbf{Adaptive top-k}: tiger forest mammal \textit{\color{blue}plant tree wild outdoor leaf mud grass rock branch daytime drink} \\ [10pt]

\multirow{3}{*}{\includegraphics[width=70pt]{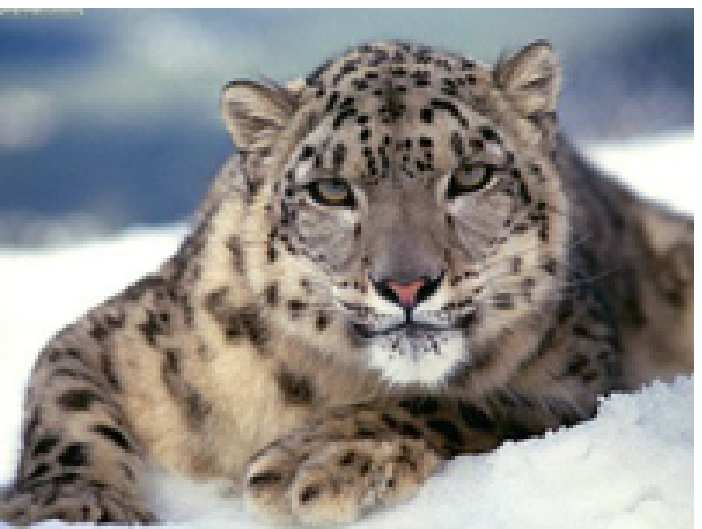}}  & image-id: H77a832fe3nqwoHq \\
& \textbf{Top-5}: cat felidae leopard wolf snow \\
& \textbf{Adaptive top-k}: cat felidae leopard wolf snow \textit{\color{blue}dog mammal tiger wild outdoor plant shadow} \\ [5pt]

Bad results: \\
\multirow{3}{*}{\includegraphics[width=70pt, height=50pt]{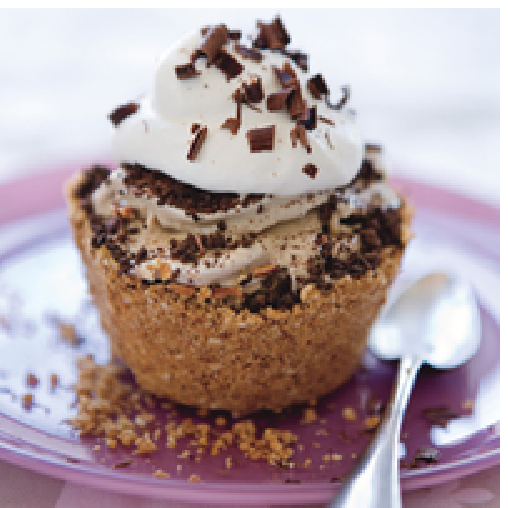}}  & image-id: x-tyO58pTX8y0hRZ \\
& \textbf{Top-5}: banana broccoli avocado food pasta \\
& \textbf{Adaptive top-k}: banana avocado \textit{\color{blue}plant tree berry fruit daytime drink rock leaf outdoor cloudless orange bus} \\ [5pt]

\multirow{3}{*}{\includegraphics[width=70pt]{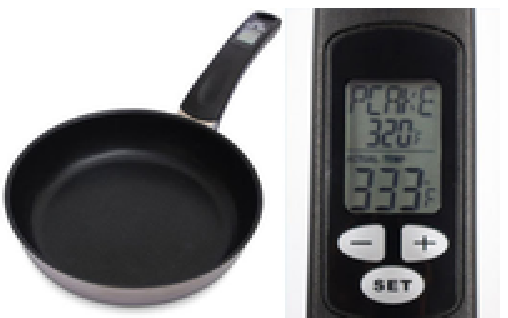}}  & image-id: zuglKl\_fHHeRoVW2 \\
& \textbf{Top-5}: phone logo bottle guitar pan \\
& \textbf{Adaptive top-k}: phone bottle \textit{\color{blue}hat drink violin indoor knife spoon shadow orange apple person human grape outdoor pear fruit bus strawberry rock leaf} \\ 

\bottomrule
\end{tabular}
}
\end{table}

\bibliographystyle{splncs}
\bibliography{clef2014}

\end{document}